\documentclass[10pt,twocolumn,letterpaper]{article}

\usepackage{iccv}
\usepackage{times}
\usepackage{epsfig}
\usepackage{graphicx}
\usepackage{amsmath}
\usepackage{amssymb}
\usepackage{xcolor}
\usepackage{wrapfig}
\usepackage{multirow}
\usepackage[pagebackref,breaklinks,colorlinks]{hyperref}

\iccvfinalcopy 

\def\iccvPaperID{5} 

\ificcvfinal\fi

\begin{document}

\title{Uni-NLX: Unifying Textual Explanations for Vision and Vision-Language Tasks}

\author{Fawaz Sammani and Nikos Deligiannis\\
ETRO Department, Vrije Universiteit Brussel, Pleinlaan 2, B-1050 Brussels, Belgium
\\imec, Kapeldreef 75, B-3001 Leuven, Belgium\\
{\tt\small fawaz.sammani@vub.be, \tt\small ndeligia@etrovub.be
}
}

\maketitle
\ificcvfinal\thispagestyle{empty}\fi

\begin{abstract}
Natural Language Explanations (NLE) aim at supplementing the prediction of a model with human-friendly natural text. Existing NLE approaches involve training separate models for each downstream task. In this work, we propose Uni-NLX, a unified framework that consolidates all NLE tasks into a single and compact multi-task model using a unified training objective of text generation. Additionally, we introduce two new NLE datasets: 1) ImageNetX, a dataset of 144K samples for explaining ImageNet categories, and 2) VQA-ParaX, a dataset of 123K samples for explaining the task of Visual Question Answering (VQA). Both datasets are derived leveraging large language models (LLMs). By training on the 1M combined NLE samples, our single unified framework is capable of simultaneously performing seven NLE tasks including VQA, visual recognition and visual reasoning tasks with 7$\times$ fewer parameters, demonstrating comparable performance to the independent task-specific models in previous approaches, and in certain tasks even outperforming them.\footnote{https://github.com/fawazsammani/uni-nlx} 

\end{abstract}

\section{Introduction}

\begin{figure}
    \centering
    \includegraphics[width=0.45\textwidth]{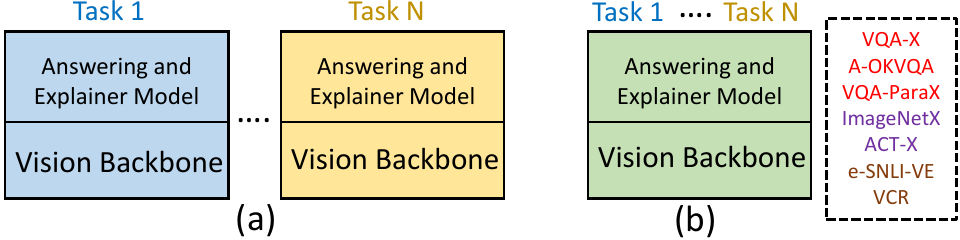}
    \caption{The current SoTA model (a) \cite{Sammani2022NLXGPTAM} unifies the answering and explainer models into a single compact model, training separate models for each of the $N$ tasks. Our proposed approach (b) takes a further step by unifying all tasks into a single compact model, resulting in $N\times$ fewer parameters. Our single unified model is capable of simultaneously handling diverse tasks ranging from \textcolor{red}{Visual Question Answering}, \textcolor{violet}{Visual Recognition} and \textcolor{brown}{Visual Reasoning}.}
    \label{demo}
\end{figure}

Moving away from general and high-level explanations such as heatmaps \cite{Selvaraju2019GradCAMVE, Sundararajan2017AxiomaticAF, Bach2015OnPE, Simonyan2014DeepIC}, Natural Language Explanations (NLE)\footnote{it is worth noting that "explanations" in this context do not refer to explanations of the underlying decision-making process of a model as typical in post-hoc explainability methods, but rather to supplementary information concerning the predicted outcome, incorporated through training} \cite{Camburu2018eSNLINL, Narang2020WT5TT} offer a detailed, human-friendly textual format explanation. Recently, NLE has been extended to encompass vision and vision-language (VL) tasks \cite{Park2018MultimodalEJ, Wu2019FaithfulME, Marasovi2020NaturalLR, Kayser2021eViLAD}. The general pipeline comprises a vision model to encode the image, a task model $M_T$ to generate a prediction for the task at hand (\textit{e.g.,} answer for VQA, class for image classification) and an explainer model $M_E$ which takes the form of a language model to produce an explanation for the prediction via natural text. A subsequent study \cite{Sammani2022NLXGPTAM} unifies $M_T$ and $M_E$ into a single compact model that performs both tasks simultaneously by converting all tasks into generative tasks with a single casual language modeling training objective (Figure \ref{demo}a). This greatly reduces the number of parameters and inference time and associates the reasoning process of $M_E$ to the same answer prediction process in $M_T$. It also attributes to the fact that explainability techniques are applied on the \textit{same} model responsible for generating the prediction. However, both these approaches require separate finetuning on each NLE task. This results in $N$ separately-parameterized models for $N$ tasks of NLE. Moreover, it requires a separate specialized model to perform each task. In this work, we build upon the work of \cite{Sammani2022NLXGPTAM} and consolidate all NLE tasks into a single compact model, dubbed as Uni-NLX (Figure~\ref{demo}b). This unification offers several advantages that previous approaches lack: Firstly, it offers a single model to simultaneously perform all $N$ NLE tasks, thereby requiring $N\times$ less parameters. Secondly, the integration enables mutual learning among all NLE tasks, as they possess similar reasoning capabilities. Lastly, the shared information across diverse tasks enables greater flexibility in answers and explanations (\textit{e.g.,} free-form text generation).

Furthermore, we propose to leverage knowledge from Large Language Models (LLMs) to obtain two additional NLE datasets: \textit{VQA-ParaX} and \textit{ImageNetX}. VQA-ParaX is a re-formulation of long-text captioning datasets (\textit{e.g.,} Image Paragraph Captioning \cite{Krause2017AHA} or Local Narratives \cite{PontTuset_eccv2020}) into question-answer-explanation formats using LLMs in a scalable manner. Moreover, LLMs posses vast knowledge about the world, and can be leveraged to obtain fine-grained, distinctive features and descriptions about different objects. ImageNetX is a dataset encompassing such textual data, which are regarded as explanations for ImageNet~\cite{Krizhevsky2012ImageNetCW} categories.   

The integration of these two additional datasets with the existing NLE datasets results in a total of 7 NLE datasets, containing approximately 1M (image, text) pairs. The textual component of these pairs comprises the question, answer, and explanation. By training on these pairs, Uni-NLX achieves performance levels comparable to state-of-the-art task-specific NLE models on 4 tasks, while surpassing them on 3 tasks. 

\section{Related Work}
Early works in NLE for vision and vision-language tasks include \cite{Hendricks2016GeneratingVE, Park2018MultimodalEJ, Li2018VQAEEE, Wu2019FaithfulME, Marasovi2020NaturalLR, Kayser2021eViLAD}. They rely on a task model (\textit{e.g.,} UNITER~\cite{Chen2020UNITERUI}) for multimodal feature extraction and answer prediction, and an explainer model (\textit{e.g.,} GPT-2~\cite{Radford2019LanguageMA}) to generate an explanation for the prediction. Most recently, NLX-GPT~\cite{Sammani2022NLXGPTAM} proposed to unify both these models into a single, compact-sized model (\textit{e.g.,} Distilled-GPT-2) that simultaneously generates and explains an answer using a single casual language modelling objective, while also eliminating the computationally-expensive object-level feature extraction stage \cite{Anderson2018BottomUpAT}. This generative formulation has also proven to be effective in vision-language pretraining methods such as VL-T5 \cite{Cho2021UnifyingVT}, OFA \cite{Wang2022UnifyingAT} and GIT \cite{Wang2022GITAG}. Multimodal-CoT \cite{Zhang2023MultimodalCR} builds upon the Chain of Thought Prompting \cite{Wei2022ChainOT} technique and instead generates a rationale (explanation) prior to generating an answer, which serves as a reasoning step for inferring the answer. However, the aforementioned methods require training or finetuning for each task individually, which consequently leads to separately-parameterized models specialized to each task. Different from these methods, our work unifies all tasks into a single compact-sized model, greatly reducing parameters and computational cost.  

The authors of \cite{Menon2022VisualCV} perform zero-shot visual classification by measuring the similarity between an image and various distinctive textual features that describe the object in the image. These descriptors are obtained from LLMs. However, this approach relies on a strong retrieval model (\textit{e.g.,} CLIP \cite{Radford2021LearningTV}) and does not have the ability to generate text. Additionally, it is primarily aimed to vision-only tasks. In contrast, our method generates flexible free-form answers and explanations for both vision and vision-language tasks.

\section{Method}
Following NLX-GPT \cite{Sammani2022NLXGPTAM}, we formulate the discriminative answer prediction task as a generative text prediction task, along with the explanation. Both the answer and explanation tasks are unified into the model which outputs a single sequence containing the answer and explanation in a textual form. We first describe how we construct additional NLE datasets, and then elaborate on our multi-task unified model. 

\subsection{Data Synthesis Strategies}
We propose to harness the powerful reasoning capabilities of LLMs to formulate two additional NLE datasets: \textit{VQA-ParaX} and \textit{ImageNetX}, in a scalable manner. We utilize GPT-3 \cite{Brown2020LanguageMA} with instructional finetuning \cite{Ouyang2022TrainingLM} (ChatGPT) as our LLM. 
\newline
\textbf{VQA-ParaX}: LLMs posses remarkable ability in reading and re-formulating passages such as summarization and information extraction. The image paragraph captioning dataset \cite{Krause2017AHA} contains 19,561 samples and provides detailed descriptions of images which allows the LLM to gain a complete understanding of the image solely through the textual description. Using a LLM, we re-formulate the image paragraph captioning dataset into question-answer-explanation formats. We prompt the LLM with \texttt{<I, \texttt{$S^i$}>}, where \texttt{$S^i$} represents the paragraph sample, and \texttt{I} represents the instruction given to the LLM. For each sample $i$, we formulate $6$ question-answer-explanation triplets, resulting in approximately 123K triplet samples. The instruction \texttt{I} we use is provided in the supplementary material.
\newline
\textbf{ImageNetX}: ImageNet-1K \cite{Krizhevsky2012ImageNetCW} is a dataset used for image classification containing 1K categories. LLMs posses wealth knowledge about the world, which can be harnessed to obtain distinctive features and descriptions of various objects at a granular level. We propose to obtain such textual descriptions from LLM for the ImageNet-1K categories, which are then regarded as explanations for the class category (answer). We prompt the LLM with \texttt{<I,c>}, where \texttt{I} represents the instruction and \texttt{c} $\in C$ represents the class category for each of the 1K categories $C$. We generate 50 descriptions for each class $c$. In order to account for variations in visual representations of the same textual description within a given class, we assign three distinct training images per description for each class. Consequently, this approach yields a dataset of approximately 141K training samples. The remaining 3K textual descriptions are associated to a single image from the ImageNet validation set, and are divided into validation and test set. The instruction \texttt{I} we use is provided in the supplementary material.

We provide further analysis, quality assessment and qualitative samples of these two new datasets in the supplementary material. 

\subsection{Unifying Explanations}
To achieve a unified NLE framework across diverse tasks, it is necessary to establish a standardized format of question-answer-explanation. However, certain tasks (\textit{e.g.,} visual recognition) lack inherent questions. To address this, we introduce a consistent question relevant to each task, such as \textit{"What category is this?"} for image recognition, \textit{"What action is this?"} for action recognition, or \textit{"is the following hypothesis true or false?"} for visual entailment. By employing this unified format, all tasks can be formulated using the sequence $S$: \texttt{<question> the answer is <answer> because <explanation>}. The compilation of all available datasets yields a collective corpus of approximately 1M samples. During training, we provide $S$ as input to the model and predict the answer and explanation component of $S$ in an autoregressive manner, utilizing a single causal language modeling training objective with cross-entropy loss. During inference, only the question is fed into the model, which subsequently predicts the answer and explanation using greedy decoding. It is worth noting that the answer can also be provided during inference, in which case the model solely generates the explanation. To allow the model to distinguish between the question, answer and explanation components of $S$, we utilize three different segment embeddings for each.

\section{Experiments}
Our unified dataset comprises seven NLE datasets encompassing visual question answering (VQA), vision recognition and visual reasoning tasks. VQA tasks consists of VQA-X \cite{Park2018MultimodalEJ} (33K samples), A-OKVQA \cite{AOKVQA} (25K samples) and VQA-ParaX (123K samples). Visual recognition tasks include ACT-X \cite{Park2018MultimodalEJ} (18K samples) for action recognition and ImageNetX (144K samples) for image classification. Visual reasoning tasks comprises e-SNLI-VE~\cite{Kayser2021eViLAD} (430K samples) for visual entailment and Visual Commensense Reasoning (VCR) \cite{Zellers2019FromRT} of 192K samples. To establish a fair comparison, our model follows NLX-GPT \cite{Sammani2022NLXGPTAM}, which uses a distilled version \cite{Sanh2019DistilBERTAD} of the GPT-2 transformer language model \cite{Brown2020LanguageMA} as the answering and explanation model, and a CLIP visual encoder part \cite{Radford2021LearningTV} as the visual backbone. Our model is trained for a maximum of 20 epochs with a batch size of 64 and a learning rate of 2e-5 which decays linearly to 0.

\begin{table}
\caption{Unfiltered Scores for Uni-NLX compared to NLX-GPT \cite{Sammani2022NLXGPTAM} on the 7 downstream tasks. Both models are w/o pretraining. B-N, M R, C, S are short for: BLEU-N, METEOR, ROUGE-L, CIDER and SPICE.}
\scalebox{0.8}{
\begin{tabular}{|c|cccccccc|}
\hline
        & \multicolumn{8}{c|}{VQA-X}                                                                                                                                                                                \\ \hline
        & \multicolumn{1}{c|}{B-1}   & \multicolumn{1}{c|}{B-2}   & \multicolumn{1}{c|}{B-3}   & \multicolumn{1}{c|}{B-4}   & \multicolumn{1}{c|}{M}    & \multicolumn{1}{c|}{R}    & \multicolumn{1}{c|}{C}     & S    \\ \hline
NLX-GPT & \multicolumn{1}{c|}{\textbf{59.1}} & \multicolumn{1}{c|}{\textbf{43.8}} & \multicolumn{1}{c|}{\textbf{32.2}} & \multicolumn{1}{c|}{\textbf{23.8}} & \multicolumn{1}{c|}{\textbf{20.3}} & \multicolumn{1}{c|}{\textbf{47.2}} & \multicolumn{1}{c|}{\textbf{89.2}}  & \textbf{18.3} \\ 
Uni-NLX & \multicolumn{1}{c|}{57.9} & \multicolumn{1}{c|}{42.1} & \multicolumn{1}{c|}{30.2} & \multicolumn{1}{c|}{21.7} & \multicolumn{1}{c|}{19.3} & \multicolumn{1}{c|}{45.9} & \multicolumn{1}{c|}{81.1}  & 17.8 \\ \hline
        & \multicolumn{8}{c|}{ACT-X}                                                                                                                                                                                \\ \hline

NLX-GPT & \multicolumn{1}{c|}{64.4} & \multicolumn{1}{c|}{47.5} & \multicolumn{1}{c|}{34.7} & \multicolumn{1}{c|}{25.6} & \multicolumn{1}{c|}{21.4} & \multicolumn{1}{c|}{48.0} & \multicolumn{1}{c|}{63.5}  & 15.4 \\ 
Uni-NLX & \multicolumn{1}{c|}{\textbf{65.4}} & \multicolumn{1}{c|}{\textbf{49.1}} & \multicolumn{1}{c|}{\textbf{36.0}} & \multicolumn{1}{c|}{\textbf{26.5}} & \multicolumn{1}{c|}{\textbf{22.0}} & \multicolumn{1}{c|}{\textbf{48.5}} & \multicolumn{1}{c|}{\textbf{67.7}}  & \textbf{16.7} \\ \hline
        & \multicolumn{8}{c|}{e-SNLI-VE}                                                                                                                                                                            \\ \hline

NLX-GPT & \multicolumn{1}{c|}{34.3} & \multicolumn{1}{c|}{22.7} & \multicolumn{1}{c|}{15.6} & \multicolumn{1}{c|}{10.9} & \multicolumn{1}{c|}{17.5} & \multicolumn{1}{c|}{31.7} & \multicolumn{1}{c|}{\textbf{106.6}} & \textbf{31.5} \\ 
Uni-NLX & \multicolumn{1}{c|}{\textbf{35.3}} & \multicolumn{1}{c|}{\textbf{23.6}} & \multicolumn{1}{c|}{\textbf{16.5}} & \multicolumn{1}{c|}{\textbf{11.8}} & \multicolumn{1}{c|}{\textbf{17.8}} & \multicolumn{1}{c|}{\textbf{32.2}} & \multicolumn{1}{c|}{106.5} & 31.3 \\ \hline
        & \multicolumn{8}{c|}{VQA-ParaX}                                                                                                                                                                            \\ \hline
NLX-GPT & \multicolumn{1}{c|}{\textbf{37.1}} & \multicolumn{1}{c|}{\textbf{27.0}} & \multicolumn{1}{c|}{\textbf{20.4}} & \multicolumn{1}{c|}{\textbf{15.5}} & \multicolumn{1}{c|}{\textbf{18.5}} & \multicolumn{1}{c|}{\textbf{40.9}} & \multicolumn{1}{c|}{\textbf{142.6}} & 31.4 \\ 
Uni-NLX & \multicolumn{1}{c|}{35.1} & \multicolumn{1}{c|}{25.7} & \multicolumn{1}{c|}{19.4} & \multicolumn{1}{c|}{14.8} & \multicolumn{1}{c|}{18.2} & \multicolumn{1}{c|}{40.8} & \multicolumn{1}{c|}{139.9} & \textbf{31.6} \\ \hline
        & \multicolumn{8}{c|}{A-OKVQA}                                                                                                                                                                              \\ \hline
NLX-GPT & \multicolumn{1}{c|}{55.0} & \multicolumn{1}{c|}{\textbf{39.9}} & \multicolumn{1}{c|}{\textbf{29.3}} & \multicolumn{1}{c|}{\textbf{20.2}} & \multicolumn{1}{c|}{16.4} & \multicolumn{1}{c|}{\textbf{46.2}} & \multicolumn{1}{c|}{\textbf{64.4}}  & 15.2 \\ 
Uni-NLX & \multicolumn{1}{c|}{\textbf{58.2}} & \multicolumn{1}{c|}{39.6} & \multicolumn{1}{c|}{27.6} & \multicolumn{1}{c|}{18.5} & \multicolumn{1}{c|}{\textbf{17.1}} & \multicolumn{1}{c|}{44.0} & \multicolumn{1}{c|}{58.1}  & \textbf{16.0} \\ \hline
        & \multicolumn{8}{c|}{ImageNetX}                                                                                                                                                                            \\ \hline
NLX-GPT & \multicolumn{1}{c|}{\textbf{64.5}} & \multicolumn{1}{c|}{\textbf{48.1}} & \multicolumn{1}{c|}{\textbf{36.9}} & \multicolumn{1}{c|}{\textbf{28.9}} & \multicolumn{1}{c|}{\textbf{22.0}} & \multicolumn{1}{c|}{\textbf{39.4}} & \multicolumn{1}{c|}{\textbf{87.5}}  & \textbf{22.4} \\ 
Uni-NLX & \multicolumn{1}{c|}{62.9} & \multicolumn{1}{c|}{46.3} & \multicolumn{1}{c|}{35.2} & \multicolumn{1}{c|}{27.4} & \multicolumn{1}{c|}{21.4} & \multicolumn{1}{c|}{38.7} & \multicolumn{1}{c|}{82.8}  & 21.3 \\ \hline
        & \multicolumn{8}{c|}{VCR}                                                                                                                                                                                  \\ \hline
NLX-GPT & \multicolumn{1}{c|}{18.5} & \multicolumn{1}{c|}{9.7}  & \multicolumn{1}{c|}{5.4}  & \multicolumn{1}{c|}{\textbf{3.3}}  & \multicolumn{1}{c|}{\textbf{9.0}}  & \multicolumn{1}{c|}{\textbf{19.9}} & \multicolumn{1}{c|}{24.2}  & 12.4 \\ 
Uni-NLX & \multicolumn{1}{c|}{\textbf{18.7}} & \multicolumn{1}{c|}{\textbf{9.9}}  & \multicolumn{1}{c|}{\textbf{5.7}}  & \multicolumn{1}{c|}{\textbf{3.5}}  & \multicolumn{1}{c|}{\textbf{9.0}}  & \multicolumn{1}{c|}{\textbf{19.9}} & \multicolumn{1}{c|}{\textbf{24.7}}  & \textbf{12.5} \\ \hline
\end{tabular}
\label{unfiltered_results}
}
\end{table}

\begin{table}
\caption{Filtered Scores for Uni-NLX compared to NLX-GPT \cite{Sammani2022NLXGPTAM} on the 7 downstream tasks. Both models are w/ pretraining.}
\scalebox{0.8}{
\begin{tabular}{|c|cccccccc|}
\hline
        & \multicolumn{8}{c|}{VQA-X}                                                                                                                                                                                                                                                        \\ \hline
        & \multicolumn{1}{c|}{B-1}            & \multicolumn{1}{c|}{B-2}            & \multicolumn{1}{c|}{B-3}            & \multicolumn{1}{c|}{B-4}            & \multicolumn{1}{c|}{M}             & \multicolumn{1}{c|}{R}             & \multicolumn{1}{c|}{C}              & S             \\ \hline
NLX-GPT & \multicolumn{1}{c|}{\textbf{64.2}} & \multicolumn{1}{c|}{\textbf{49.5}} & \multicolumn{1}{c|}{\textbf{37.6}} & \multicolumn{1}{c|}{\textbf{28.5}} & \multicolumn{1}{c|}{\textbf{23.1}} & \multicolumn{1}{c|}{\textbf{51.5}} & \multicolumn{1}{c|}{\textbf{110.6}} & \textbf{22.1} \\ 
Uni-NLX & \multicolumn{1}{c|}{62.1}          & \multicolumn{1}{c|}{46.8}          & \multicolumn{1}{c|}{34.9}          & \multicolumn{1}{c|}{26.0}          & \multicolumn{1}{c|}{21.8}          & \multicolumn{1}{c|}{48.8}          & \multicolumn{1}{c|}{97.8}           & 20.8          \\ \hline
        & \multicolumn{8}{c|}{ACT-X}                                                                                                                                                                                                                                                        \\ \hline
        & \multicolumn{1}{c|}{B1}            & \multicolumn{1}{c|}{B2}            & \multicolumn{1}{c|}{B3}            & \multicolumn{1}{c|}{B4}            & \multicolumn{1}{c|}{M}             & \multicolumn{1}{c|}{R}             & \multicolumn{1}{c|}{C}              & S             \\ \hline
NLX-GPT & \multicolumn{1}{c|}{\textbf{71.6}} & \multicolumn{1}{c|}{56.2}          & \multicolumn{1}{c|}{43.2}          & \multicolumn{1}{c|}{\textbf{33.5}} & \multicolumn{1}{c|}{\textbf{25.7}} & \multicolumn{1}{c|}{\textbf{53.7}} & \multicolumn{1}{c|}{\textbf{111.8}} & \textbf{23.3} \\
Uni-NLX & \multicolumn{1}{c|}{71.5}          & \multicolumn{1}{c|}{\textbf{56.7}} & \multicolumn{1}{c|}{\textbf{43.6}} & \multicolumn{1}{c|}{\textbf{33.5}} & \multicolumn{1}{c|}{\textbf{25.7}} & \multicolumn{1}{c|}{53.5}          & \multicolumn{1}{c|}{109.4}          & 22.8          \\ \hline
        & \multicolumn{8}{c|}{e-SNLI-VE}                                                                                                                                                                                                                                                    \\ \hline
        & \multicolumn{1}{c|}{B1}            & \multicolumn{1}{c|}{B2}            & \multicolumn{1}{c|}{B3}            & \multicolumn{1}{c|}{B4}            & \multicolumn{1}{c|}{M}             & \multicolumn{1}{c|}{R}             & \multicolumn{1}{c|}{C}              & S             \\ \hline
NLX-GPT & \multicolumn{1}{c|}{\textbf{35.7}} & \multicolumn{1}{c|}{24.0}          & \multicolumn{1}{c|}{16.8}          & \multicolumn{1}{c|}{11.9}          & \multicolumn{1}{c|}{18.1}          & \multicolumn{1}{c|}{33.4}          & \multicolumn{1}{c|}{114.7}          & \textbf{32.1} \\ 
Uni-NLX & \multicolumn{1}{c|}{35.3}          & \multicolumn{1}{c|}{\textbf{24.1}} & \multicolumn{1}{c|}{\textbf{17.0}} & \multicolumn{1}{c|}{\textbf{12.3}} & \multicolumn{1}{c|}{\textbf{18.2}} & \multicolumn{1}{c|}{\textbf{33.7}} & \multicolumn{1}{c|}{\textbf{115.4}} & \textbf{32.1} \\ \hline
        & \multicolumn{8}{c|}{VQA-ParaX}                                                                                                                                                                                                                                                    \\ \hline
        & \multicolumn{1}{c|}{B1}            & \multicolumn{1}{c|}{B2}            & \multicolumn{1}{c|}{B3}            & \multicolumn{1}{c|}{B4}            & \multicolumn{1}{c|}{M}             & \multicolumn{1}{c|}{R}             & \multicolumn{1}{c|}{C}              & S             \\ \hline
NLX-GPT & \multicolumn{1}{c|}{\textbf{41.9}} & \multicolumn{1}{c|}{\textbf{31.5}} & \multicolumn{1}{c|}{\textbf{24.7}} & \multicolumn{1}{c|}{\textbf{19.9}} & \multicolumn{1}{c|}{\textbf{22.3}} & \multicolumn{1}{c|}{\textbf{47.2}} & \multicolumn{1}{c|}{\textbf{203.7}} & 41.9          \\ 
Uni-NLX & \multicolumn{1}{c|}{41.3}          & \multicolumn{1}{c|}{31.2}          & \multicolumn{1}{c|}{24.5}          & \multicolumn{1}{c|}{19.7}          & \multicolumn{1}{c|}{22.0}          & \multicolumn{1}{c|}{\textbf{47.2}} & \multicolumn{1}{c|}{203.6}          & \textbf{42.1} \\ \hline
        & \multicolumn{8}{c|}{A-OKVQA}                                                                                                                                                                                                                                                      \\ \hline
        & \multicolumn{1}{c|}{B1}            & \multicolumn{1}{c|}{B2}            & \multicolumn{1}{c|}{B3}            & \multicolumn{1}{c|}{B4}            & \multicolumn{1}{c|}{M}             & \multicolumn{1}{c|}{R}             & \multicolumn{1}{c|}{C}              & S             \\ \hline
NLX-GPT & \multicolumn{1}{c|}{\textbf{62.3}} & \multicolumn{1}{c|}{\textbf{46.8}} & \multicolumn{1}{c|}{\textbf{36.1}} & \multicolumn{1}{c|}{\textbf{27.7}} & \multicolumn{1}{c|}{\textbf{20.5}} & \multicolumn{1}{c|}{\textbf{51.5}} & \multicolumn{1}{c|}{\textbf{93.0}}  & 19.3          \\ 
Uni-NLX & \multicolumn{1}{c|}{62.1}          & \multicolumn{1}{c|}{43.3}          & \multicolumn{1}{c|}{30.8}          & \multicolumn{1}{c|}{20.8}          & \multicolumn{1}{c|}{19.6}          & \multicolumn{1}{c|}{48.1}          & \multicolumn{1}{c|}{78.1}           & \textbf{19.7} \\ \hline
        & \multicolumn{8}{c|}{ImageNetX}                                                                                                                                                                                                                                                    \\ \hline
        & \multicolumn{1}{c|}{B1}            & \multicolumn{1}{c|}{B2}            & \multicolumn{1}{c|}{B3}            & \multicolumn{1}{c|}{B4}            & \multicolumn{1}{c|}{M}             & \multicolumn{1}{c|}{R}             & \multicolumn{1}{c|}{C}              & S             \\ \hline
NLX-GPT & \multicolumn{1}{c|}{69.7}          & \multicolumn{1}{c|}{54.1}          & \multicolumn{1}{c|}{42.5}          & \multicolumn{1}{c|}{33.8}          & \multicolumn{1}{c|}{24.7}          & \multicolumn{1}{c|}{43.1}          & \multicolumn{1}{c|}{107.4}          & 26.1          \\ 
Uni-NLX & \multicolumn{1}{c|}{\textbf{71.9}} & \multicolumn{1}{c|}{\textbf{56.5}} & \multicolumn{1}{c|}{\textbf{45.0}} & \multicolumn{1}{c|}{\textbf{36.1}} & \multicolumn{1}{c|}{\textbf{25.8}} & \multicolumn{1}{c|}{\textbf{44.8}} & \multicolumn{1}{c|}{\textbf{117.2}} & \textbf{27.3} \\ \hline
        & \multicolumn{8}{c|}{VCR}                                                                                                                                                                                                                                                          \\ \hline
        & \multicolumn{1}{c|}{B1}            & \multicolumn{1}{c|}{B2}            & \multicolumn{1}{c|}{B3}            & \multicolumn{1}{c|}{B4}            & \multicolumn{1}{c|}{M}             & \multicolumn{1}{c|}{R}             & \multicolumn{1}{c|}{C}              & S             \\ \hline
NLX-GPT & \multicolumn{1}{c|}{-}             & \multicolumn{1}{c|}{-}             & \multicolumn{1}{c|}{-}             & \multicolumn{1}{c|}{-}             & \multicolumn{1}{c|}{-}             & \multicolumn{1}{c|}{-}             & \multicolumn{1}{c|}{-}              & -             \\ 
Uni-NLX & \multicolumn{1}{c|}{29.7}          & \multicolumn{1}{c|}{23.4}          & \multicolumn{1}{c|}{19.9}          & \multicolumn{1}{c|}{17.4}          & \multicolumn{1}{c|}{17.1}          & \multicolumn{1}{c|}{33.6}          & \multicolumn{1}{c|}{85.7}           & 23.5          \\ \hline
\end{tabular}
\label{filtered_results}
}
\end{table}

\begin{figure*}
    \centering
    \includegraphics[width=0.87\textwidth]{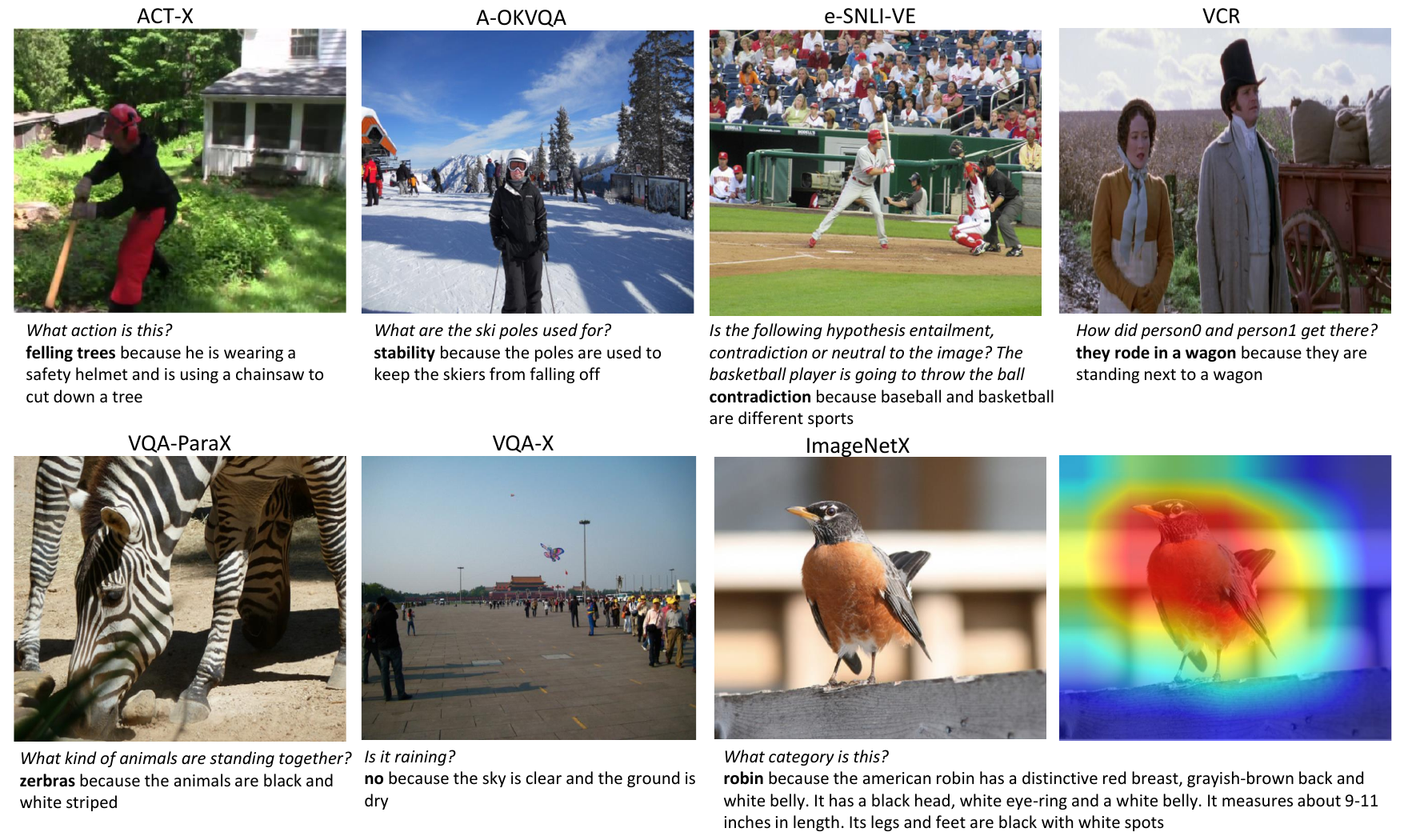}
    \caption{Qualitative Examples of Uni-NLX on the 7 NLE tasks. We show the \textit{question}, \textbf{answer} and explanation under each image.}
    \label{qualitative}
\end{figure*}

\subsection{Quantitative Results}
We evaluate our model quantitatively using automatic natural language generation (NLG) metrics (BLEU~\cite{Papineni2001BleuAM}, METEOR~\cite{Banerjee2005METEORAA}, ROUGE-L~\cite{Lin2004ROUGEAP}, CIDER~\cite{Vedantam2014CIDErCI} and SPICE~\cite{Anderson2016SPICESP}); all scores are computed with the publicly available code\footnote{https://github.com/tylin/coco-caption}. Following previous works, the evaluation is carried on in two settings: \textit{filtered} and \textit{unfiltered}. In the filtered setting, we only consider the explanations for which the predicted answer is correct. In the unfiltered setting, all explanations are considered, irrespective of whether the predicted answer associated with each explanation is true or false. We utilize the recent state-of-the-art NLX-GPT \cite{Sammani2022NLXGPTAM} model as our baseline for evaluating our approach. NLX-GPT also presents results of NLE tasks by fine-tuning a pretrained model on image captioning. In our study, we consider this setting and utilize the pretrained model provided by the official code\footnote{https://github.com/fawazsammani/nlxgpt}. Table \ref{unfiltered_results} presents the unfiltered results of Uni-NLX without finetuning the pretrained model, while Table \ref{filtered_results} reports the filtered results obtained after finetuning the pretrained model. Additional results on unfiltered results with pretraining and filtered results without pretraining can be found in the supplementary material. In Table \ref{unfiltered_results}, Uni-NLX demonstrates superior performance compared to NLX-GPT on ACT-X, e-SNLI-VE, and VCR. Additionally, Uni-NLX achieves performance that is comparable to NLX-GPT across all VQA tasks (VQA-X, VQA-ParaX, and A-OKVQA) and ImageNetX, and surpasses NLX-GPT on certain metrics. Table \ref{filtered_results} shows that Uni-NLX outperforms NLX-GPT on e-SNLI-VE and ImageNetX and demonstrates comparable performance to other tasks, and in certain metrics even outperforms them. It is worth noting that NLX-GPT does not present unfiltered results on VCR. 

\subsection{Qualitative Results}
Figure \ref{qualitative} shows qualitative results for each of the seven NLE tasks. As observed, our model generates an answer to the given question and image, supported by a detailed explanation. We discuss limitations such as collapse cases in the supplementary material. For ImageNetX, we additionally show a heatmap visualization obtained from ResNet-18 \cite{He2016DeepRL} using Grad-CAM \cite{Selvaraju2019GradCAMVE}, which only displays high-level and general features influencing the prediction. On the other hand, Uni-NLX provides distinctive and fine-grained concepts which influenced the prediction (\textit{e.g.,} red breast, grayish-brown back, black with white spots) in the form of human-friendly text. 
\begin{wraptable}{r}{3cm}
\caption{CLIP Zero-Shot Transfer Results}
\scalebox{0.8}{
\begin{tabular}{|c|c|}
\hline
        & Top-1 (\%)    \\ \hline
CLIP    & 64.9          \\ \hline
Uni-NLX & \textbf{66.6} \\ \hline
\end{tabular}
}
\label{clip_zero_shot}
\end{wraptable} 
Furthermore, the attribution maps associated with these distinctive textual attributes have the potential to represent automatically-extracted prototypes. In Table \ref{clip_zero_shot}, the generated ImageNetX explanations yields a noteworthy improvement of 1.7\% on ImageNet CLIP zero-shot transfer performance with ViT-B/16, without requiring a LLM as in previous works \cite{Menon2022VisualCV, Pratt2022WhatDA}. More information is provided in the supplementary material. 

\section{Conclusion}
We proposed Uni-NLX, a unified model which simultaneously performs seven NLE tasks. Leveraging a LLM, we also introduced two additional NLE datasets: VQA-Parax for the VQA task, and ImageNetX for the ImageNet recognition task. Experiments demonstrate that Uni-NLX achieves comparable performance to task-specific models in certain tasks, while surpassing them on others. 

{\small
\bibliographystyle{ieee_fullname}
\bibliography{egbib}
}

\clearpage
{\Large \textbf{Supplementary Material}}
\newline

\setcounter{section}{0}
\setcounter{figure}{0}
\setcounter{table}{0}

\section{Prompts}
In this section, we provide the prompts we use to formulate the VQA-ParaX and ImageNetX Natural Language Explanation (NLE) datasets.

\paragraph{VQA-ParaX}: We prompt the Large Language Model (LLM) with \texttt{<I, \texttt{$S^i$}>}. This consists of the paragraph sample \texttt{$S^i$} and the instruction \texttt{I}. The instruction \texttt{I} is constructed with the following considerations: 
\begin{itemize}
  \item An overview of the task that the LLM has to perform
  \item A guideline to generate a short answer, typical of the standard Visual Question Answering (VQA) scenario
  \item A guideline to avoid trivial cases
  \item An example of the task and of the trivial case, following the few-shot learning paradigm.
  \item An output format to facilitate the post-processing stage.
\end{itemize}

\texttt{I} is formulated as follows: \texttt{You are an assistant which helps formulate a VQA dataset with Textual Explanations to train deep learning models. Read the following text and formulate 3 samples, as unique as possible, each consisting of a question (Q), answer (A) and more information about the answer to help in better understanding it (E). The answers should be short, maximum of 3 words. Here is an example for Q, A and E, respectively: Q: What sport is being played?, A: baseball, E: they are playing on a baseball diamond with a ball and a bat. Also, E should be non-trivial. For example, if Q is: Where is the green tennis ball? and A is: above her head, then E should NOT BE: there is a green tennis ball above the woman's head. This is considered as trivial. Please generate the output in a single line strictly following this format for the 3 samples, where <r> indicates your response: [{Q:<r>, A:<r>, E:<r>}, {Q:<r>, A:<r>, E:<r>}, {Q:<r>, A:<r>, E:<r>}]. Here is the text:}

It is worth noting that in preliminary stages of this work, the sub-instruction: \texttt{and more information about the answer to help in better understanding it (E)} was formulated as \texttt{and an explanation (E) to explain the answer (A)}. However, we observed that in the majority of cases, the outputs from the LLM were primarily focused on its own reasoning process, consistently generating trivial statements: "the text describes that..." or "it is mentioned in the text that...". Consequently, we decided to avoid this particular sub-instruction from further consideration.

\paragraph{ImageNetX}: We prompt the LLM with \texttt{<I,c>}, where \texttt{I} represents the instruction and \texttt{c} represents the class category. The instruction \texttt{I} 
consists of the following:

\begin{itemize}
  \item An overview of the task that the LLM has to perform
  \item An additional guideline to tune the generated output to be general and short, avoiding an extensive enumeration of individual elements pertaining to class $c$
\end{itemize}

The instruction \texttt{I} is formulated as: \texttt{You are an assistant which helps humans describe objects. what are physical features and characteristics describing a $c$? Please answer in a short, brief and concise way, with a maximum of 50 words.}

\begin{figure*}
    \centering
    \includegraphics[width=\textwidth]{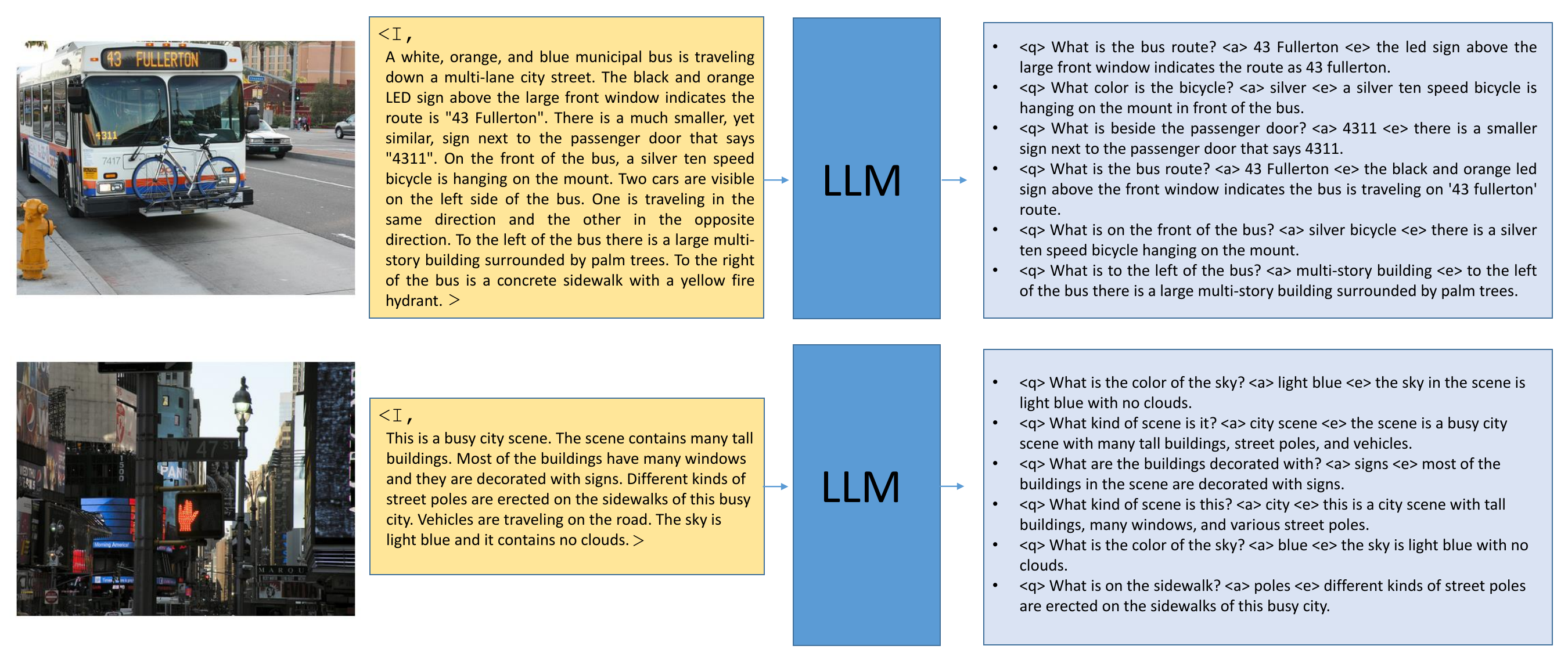}
    \caption{The process of generating VQA-ParaX leveraging a Large Language Model (LLM). The instruction \texttt{I} and the text fragment describing the image jointly serve as the input prompt (yellow box) for the LLM, which reformulates the text fragment into 6 samples, each consisting of a question $<q>$, answer $<a>$ and explanation $<e>$. It is important to note that the image is not provided to the LLM.}
    \label{qualitative samples_vqaparaX}
\end{figure*}

\begin{figure*}
    \centering
    \includegraphics[width=\textwidth]{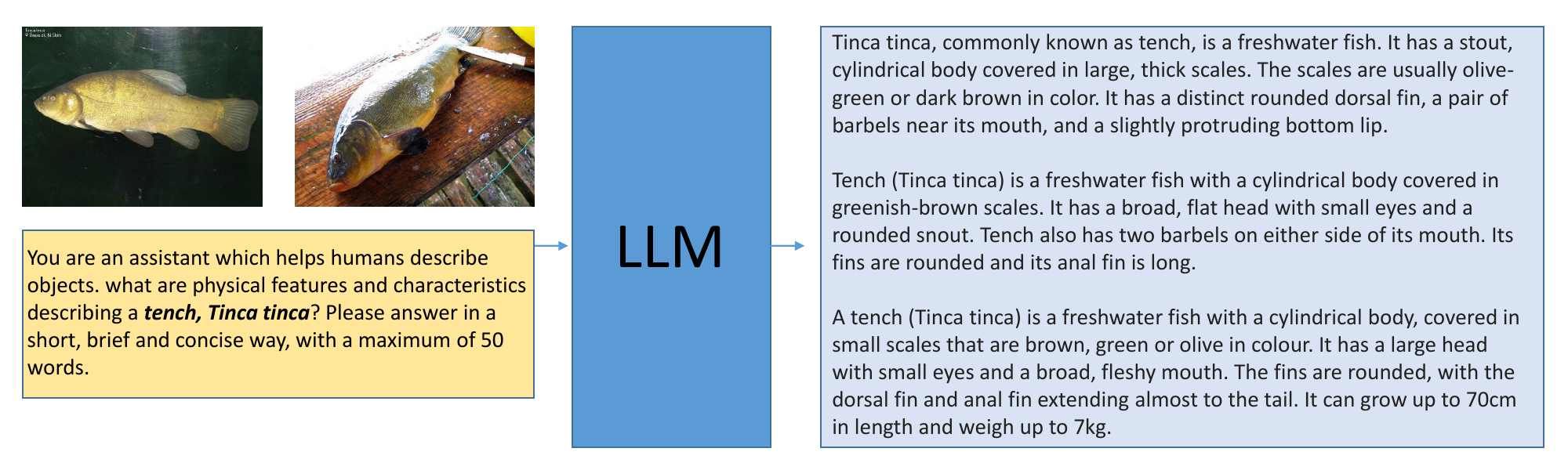}
    \caption{The process of generating ImageNetX leveraging a LLM. The instruction containing the ImageNet category (\textit{tench} in this example) serves as the input prompt (yellow box) for the LLM, which outputs distinctive features describing that category. Although 3 generated samples are shown, it is important to clarify that we generate a single sample for each run. It is important to also note that no image is provided to the LLM.}
    \label{qualitative samples_imagenetX}
\end{figure*}

\begin{figure*}
    \centering
    \includegraphics[width=\textwidth]{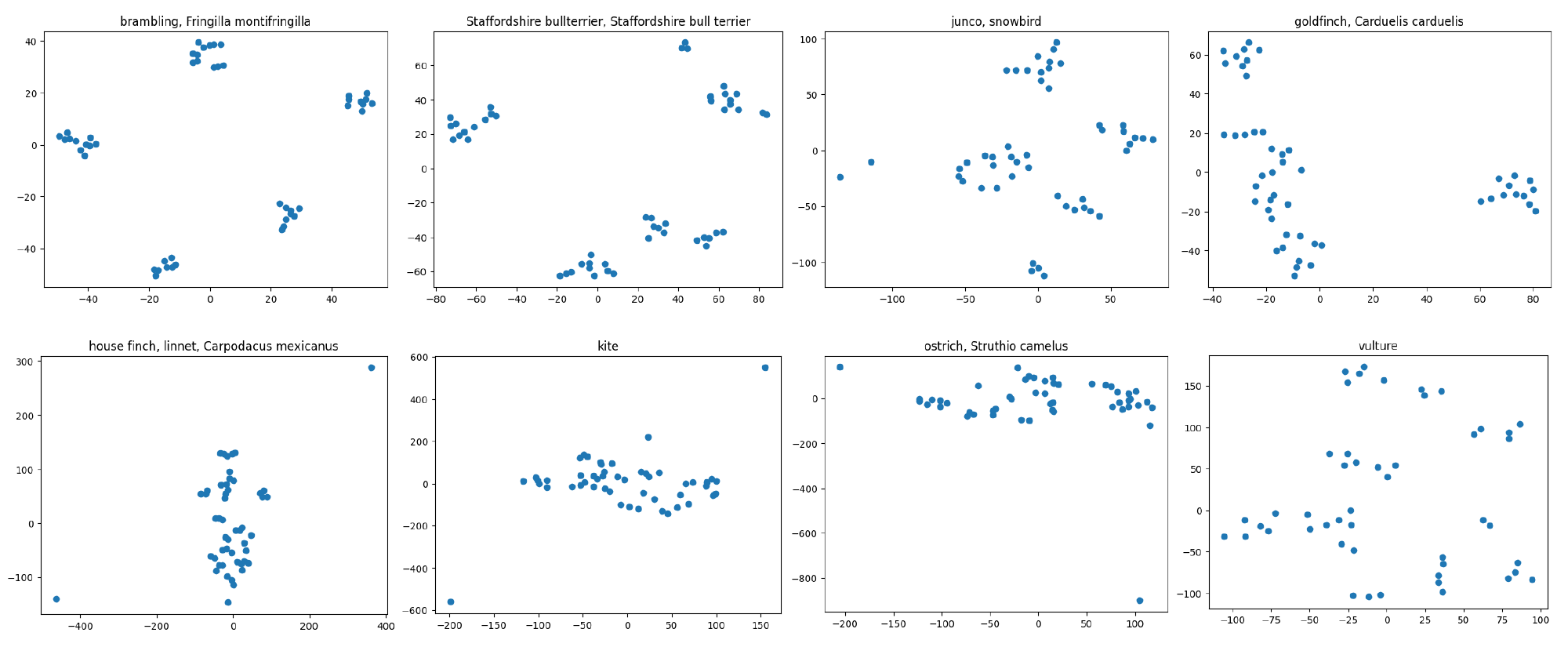}
    \caption{A 2D visualization using t-SNE \cite{Maaten2008VisualizingDU} of the different textual descriptions for ImageNet categories generated using the LLM. Each plot depicts a distinct ImageNet category, with each data point representing 1 of 50 sample descriptions produced for each category.}  
    \label{imagenet_tsne}
\end{figure*}

\section{Qualitative Data Samples}
Figure \ref{qualitative samples_vqaparaX} shows 2 examples depicting the process of generating VQA-ParaX. The instruction \texttt{I} and the text fragment describing the image jointly serve as the input prompt for the LLM. The output of the LLM is a re-formulation of the text fragment into 6 samples, each consisting of a question $<q>$, answer $<a>$ and explanation $<e>$.  Similarly, Figure \ref{qualitative samples_imagenetX} shows an example illustrating the process of generating ImageNetX. The instruction containing the ImageNet category serves as the input prompt for the LLM. The output is a textual description describing attributes and distinctive features of that category. 

\section{Data Analysis and Quality Assessment}
In this section, we perform analysis on the newly introduced datasets VQA-ParaX and ImageNetX. Subsequently, we evaluate their quality through ablation experiments. Table \ref{word_len} presents the average and maximum word lengths of the explanations for both VQA-ParaX and ImageNetX. As observed, the average word length of VQA-ParaX explanations is similar to the average word length of explanations from other NLE datasets. ImageNetX has a larger average word length describing distinctive features requires more words. In Table \ref{repetition}, we present question repetition statistics on VQA-ParaX. As the prompt requests 6 samples, the LLM might reiterate the constructed questions when the provided textual description of the image is overly brief and lacks information (\textit{e.g.,} a textual description as: \textit{a man in a white t-shirt and blue jeans and a cell phone}). However, even though the question and answer might be replicated across a sample, the explanation is typically formulated differently.  

Next, we examine the uniqueness of the 50 different samples we generate for each ImageNet category. We first encode each sample through MPNet \cite{Song2020MPNetMA} finetuned on 1B sentence pairs using the self-supervised contrastive learning objective, utilizing the Sentence-Transformers \cite{reimers-2019-sentence-bert} library\footnote{https://github.com/UKPLab/sentence-transformers} to obtain a 768-d vector representing the sample description. We apply t-SNE \cite{Maaten2008VisualizingDU} to reduce the vector into a 2-d space for visualization. Upon plotting the 50 samples, we observe distinct clusters emerging among them, highlighting the diversity present across the samples, as illustrated in the upper row. Conversely, the initial three instances depicted in the lower row reveal that these samples tend to exhibit greater similarity, resulting in the formation of a singular cluster.

Lastly, we perform ablation studies on the newly introduced datasets VQA-ParaX and ImageNetX in Table \ref{data_ablation_restructured}. We start by analyzing the exclusion of VQA-ParaX from the training NLE corpus. In 60\% of the cases (across all datasets and metrics), the inclusion of VQA-ParaX improves the evaluation metrics across the board. This suggests that VQA-ParaX contribute positively to the performance of the model. Next, we evaluate the exclusion of ImageNetX. In 94\% of the cases, excluding ImageNetX improves the performance of evaluation metrics across all VQA NLE tasks (A-OKVQA, VQA-X and VQA-ParaX), and including it improves the performance of visual recognition and visual reasoning tasks in 84\% of the cases. This suggests that ImageNetX has a negative impact on VQA tasks, but positive impact on the other tasks. This can be rationalized by considering the complexity of ImageNetX, which requires the model to additionally learn how to describe coarse-grained distinctive textual features of an object. This task is more challenging compared to VQA tasks. As a result, incorporating a broader range of such complex information might lead to a trade-off, resulting in a decrease in performance for VQA tasks. Finally, we investigate the performance of the model when excluding both VQA-ParaX and ImageNetX. We find that in 58\% of the cases, excluding both these datasets leads to an improvement in performance. 

\begin{table}[]
\centering
\caption{Average Word Length of VQA-ParaX and ImageNetX}
\begin{tabular}{|c|c|c|}
\hline
                    & VQA-ParaX & ImageNetX \\ \hline
Average Word Length & 13        & 49        \\ 
Maximum Word Length & 90        & 110       \\ \hline
\end{tabular}
\label{word_len}
\end{table}

\begin{table}[]
\centering
\caption{Question Repetition statistics for VQA-ParaX}
\begin{tabular}{|c|c|}
\hline
\textbf{Repetitions}                  & \textbf{Value}   \\ \hline
Maximum repetitions across all samples     & 3       \\ 
Percentage of samples with 3 repetitions & 1.96\%  \\
Percentage of samples with 2 repetitions   & 12.96\% \\ 
Percentage of samples with 1 repetition   & 36.84\% \\ 
Percentage of samples with no repetitions  & 48.23\% \\ \hline
\end{tabular}
\label{repetition}
\end{table}

\begin{table*}
\centering
\caption{Ablation Studies on the newly introduced datasets (Unfiltered Scores, w/o pretraining). B-N, M R, C, S are short for: BLEU-N, METEOR, ROUGE-L, CIDER and SPICE.}

\begin{tabular}{|c|c|c|c|c|c|c|c|}
\hline
\textbf{Dataset} & \textbf{Setting} & \textbf{B1} & \textbf{B4} & \textbf{M} & \textbf{R} & \textbf{C} & \textbf{S} \\ \hline
\multirow{4}{*}{ACT-X} & All Data (Uni-NLX) & 0.654 & 0.265 & \textbf{0.220} & 0.485 & 0.677 & \textbf{0.167} \\
 & w/o VQA-ParaX & \textbf{0.658} & 0.265 & 0.219 & 0.484 & 0.680 & 0.166 \\
 & w/o ImageNetX & 0.656 & 0.263 & 0.219 & 0.483 & 0.675 & \textbf{0.167} \\
 & w/o VQA-ParaX, ImageNetX & 0.655 & \textbf{0.271} & 0.219 & \textbf{0.486} & \textbf{0.685} & 0.165 \\ \hline
\multirow{4}{*}{A-OKVQA} & All Data (Uni-NLX) & \textbf{0.582} & 0.185 & 0.171 & 0.440 & 0.581 & 0.160 \\
 & w/o VQA-ParaX & 0.561 & \textbf{0.209} & 0.168 & \textbf{0.458} & \textbf{0.652} & 0.152 \\
 & w/o ImageNetX & 0.576 & 0.194 & \textbf{0.173} & 0.445 & 0.608 & \textbf{0.161} \\
 & w/o VQA-ParaX, ImageNetX & 0.558 & 0.198 & 0.166 & 0.455 & 0.624 & 0.155 \\ \hline
\multirow{4}{*}{VQA-X} & All Data (Uni-NLX) & 0.579 & 0.217 & 0.194 & 0.459 & 0.811 & 0.178 \\
 & w/o VQA-ParaX & 0.578 & 0.224 & 0.196 & 0.463 & 0.833 & 0.176 \\
 & w/o ImageNetX & \textbf{0.588} & \textbf{0.232} & \textbf{0.202} & \textbf{0.472} & \textbf{0.865} & \textbf{0.182} \\
 & w/o VQA-ParaX, ImageNetX & 0.578 & 0.221 & 0.196 & 0.462 & 0.818 & 0.179 \\ \hline
\multirow{4}{*}{VQA-ParaX} & All Data (Uni-NLX) & 0.351 & 0.148 & 0.182 & 0.408 & 1.399 & \textbf{0.316} \\
 & w/o VQA-ParaX & 0.165 & 0.058 & 0.120 & 0.319 & 0.769 & 0.230 \\
 & w/o ImageNetX & \textbf{0.360} & \textbf{0.151} & \textbf{0.183} & \textbf{0.409} & \textbf{1.416} & \textbf{0.316} \\
 & w/o VQA-ParaX, ImageNetX & 0.164 & 0.058 & 0.118 & 0.317 & 0.758 & 0.226 \\ \hline
\multirow{4}{*}{e-SNLI-VE} & All Data (Uni-NLX) & \textbf{0.353} & \textbf{0.118} & \textbf{0.178} & \textbf{0.322} & 1.065 & 0.313 \\
 & w/o VQA-ParaX & 0.343 & 0.112 & 0.174 & 0.318 & 1.044 & 0.308 \\
 & w/o ImageNetX & 0.327 & 0.104 & 0.169 & 0.311 & 1.023 & 0.307 \\
 & w/o VQA-ParaX, ImageNetX & 0.348 & 0.115 & 0.177 & 0.321 & \textbf{1.066} & \textbf{0.319} \\ \hline
\end{tabular}

\label{data_ablation_restructured}
\end{table*}

\section{Additional Qualitative Examples}
\begin{figure*}
    \centering
    \includegraphics[width=\textwidth]{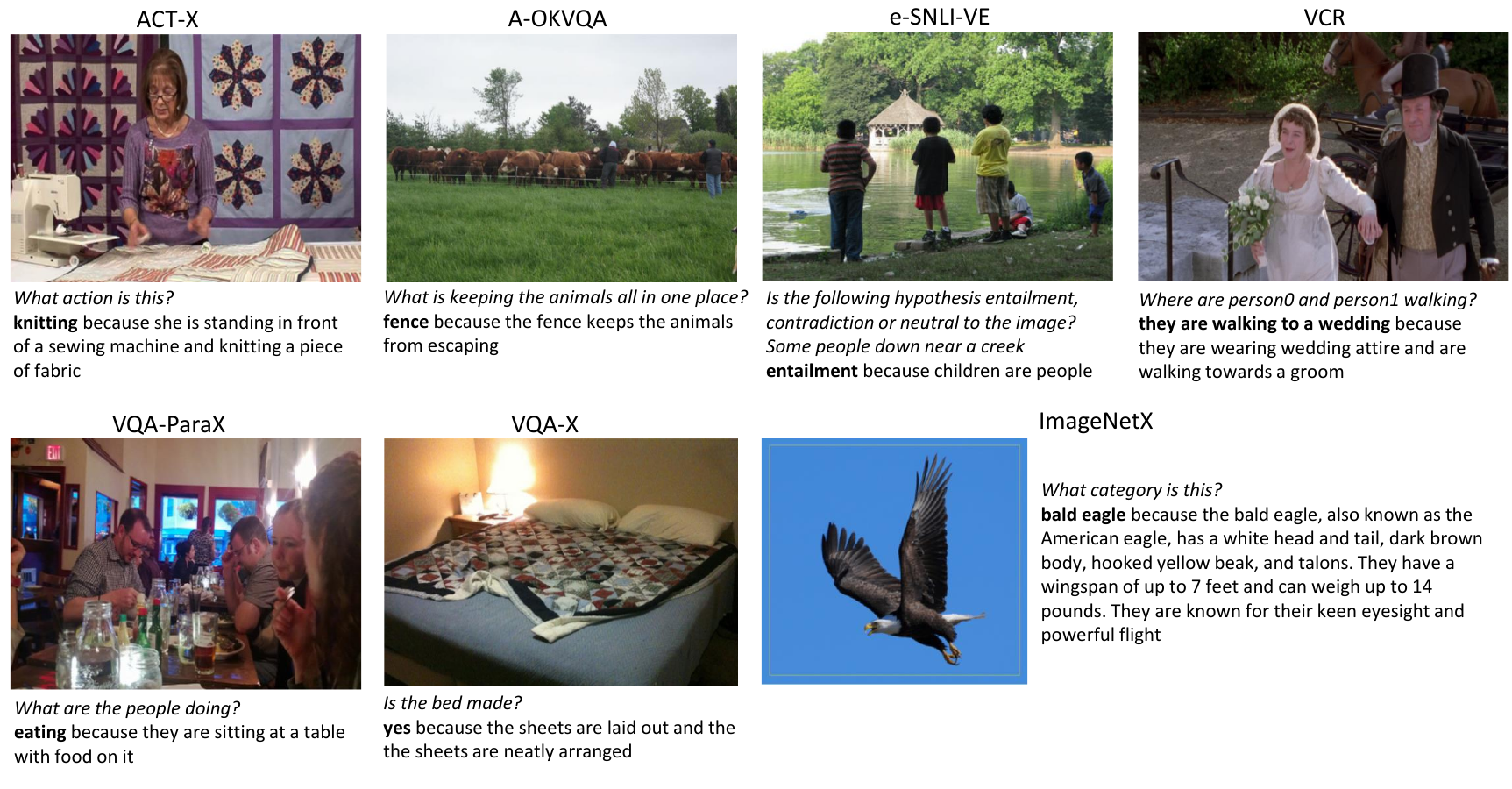}
    \caption{Qualitative Examples of Uni-NLX on the 7 NLE tasks. We show the \textit{question}, \textbf{answer} and explanation under each image.}
    \label{qualitativesupp}
\end{figure*}

\begin{figure*}
    \centering
    \includegraphics[width=\textwidth]{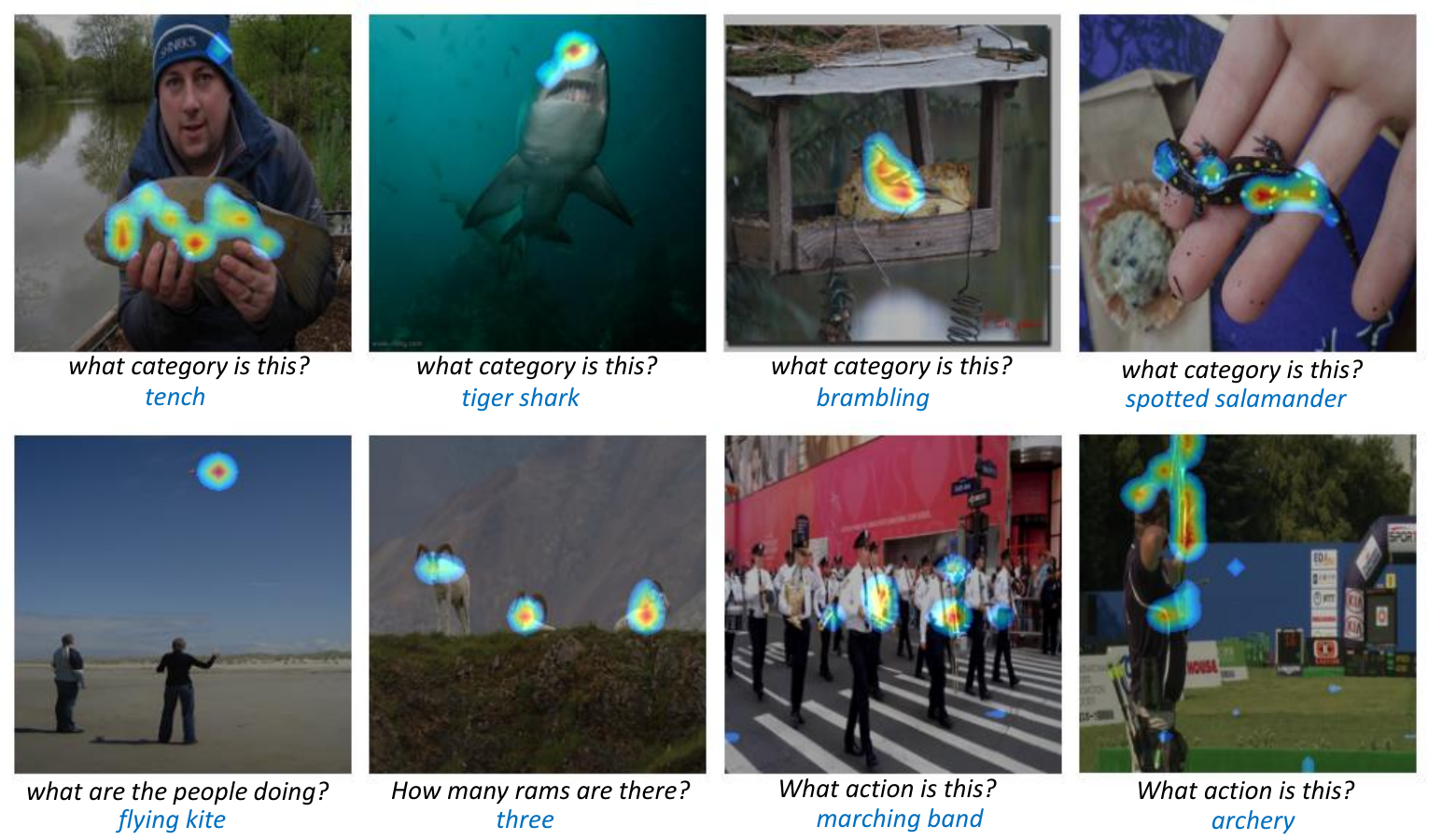}
    \caption{The attention maps for the generated answers of ImageNetX (top row), VQA-ParaX (bottom-left) and ACT-X (bottom-right).}
    \label{heatmap_answers}
\end{figure*}

We provide additional qualitative examples for each of the seven NLE tasks in Figure \ref{qualitativesupp}. As evidenced in our observations, our model generates an answer to the provided question about a given image, complemented by an explanation. For ImageNetX, it becomes apparent that Uni-NLX offers distinctive, fine-grained explanations for the predicted answer (\textit{e.g.,} white head and tail, dark brown body, yellow beak, large wingspan, the weight), all conveyed in a manner easily understandable to humans. In Figure \ref{heatmap_answers}, we visualize the attention maps for the generated answers from the last layer of the model. We analyze ImageNetX answers in the top row, VQA-ParaX answers in the bottom-left and ACT-X answers in the bottom-right. As demonstrated for ImageNetX, the presented heatmaps exhibit distinctive features within the image, in contrast to conventional explainability techniques that usually yield heatmaps encompassing the entire main object in the image.

\section{CLIP Zero-Shot Transfer Results}
We investigate CLIP zero-shot transfer results on the ImageNet ILSVRC2012 validation set. Similar to \cite{Radford2021LearningTV}, we transform each class category into a textual prompt and subsequently encode all class prompts with the CLIP text decoder. We then encode each validation image using the CLIP vision encoder. Since both models embed the image and text into a shared representation space, the class with the highest similarity score in this space is considered as the predicted class. 

In zero-shot transfer, we assume that the class categories are readily available. We utilize NLX-GPT trained on ImageNetX to generate an explanation for a class we condition on. The process involves providing our distilled GPT-2 model with the prompt "the answer is \{class label\} because" and subsequently obtaining the generated explanation. We select a random image for each class category and perform this procedure. We then formulate each \{class label\} and its generated \{explanation\} as a textual prompt, and encode all these prompts using the CLIP text decoder. We then we encode each validation image using the CLIP vision encoder and take the class with the highest similarity score in the feature unified vision-language space as the predicted class. 

Table \ref{prompt_zero_shot_transfer} shows our top-1 accuracy results across different prompts, comparing the performance when only the \{class label\} is provided in the prompt, and when both the \{class label\} and \{explanation\} are provided in the prompt. Notably, when an explanation is provided, the prompt "how can you identify" achieves the highest performance at 65.5\%, while its effectiveness decreases to 61.7\% when no explanation is provided in the prompt. Additionally, incorporating the sub-prompt "Distinctive and physical features describing" before the explanation results in an improved top-1 accuracy of 66.0\%. Due to variations in the generated explanations for different images, we adopt an approach of averaging their textual representations before calculating similarity with the image features. This approach leads to an incremental improvement in top-1 accuracy: 0.3\% when averaging two explanations per category, 0.5\% for three explanations per category, and 0.6\% for four explanations per category.

Prior work \cite{Menon2022VisualCV, Pratt2022WhatDA} improve zero-shot transfer accuracy by utilizing textual attributes that describe class categories. These attributes are generated with a Large Language Model (LLM). However, such approaches necessitate the deployment of a LLM with a substantial parameter count (at least 7B parameters). In contrast, our approach harnesses a distilled GPT-2 model, considerably smaller in size ($\sim$93M parameters), to produce explanations. These explanations are characterized by attributes, fine-grained specificity, and distinctiveness of class categories. This obviates the requirement for a resource-intensive LLM for zero-shot transfer.

\begin{table*}[h]
\centering
\caption{CLIP Zero-Shot Transfer Results on ImageNet with ViT-B/16 for different prompts, using the ImageNetX generated \{explanation\}. Since different images yield different generated explanations (E), we may average their textual representations prior to calculating the similarity with the image features.}
\begin{tabular}{|cc|}
\hline
\multicolumn{1}{|c|}{\textbf{Prompt}}                                                                                                       & \textbf{Top-1 (\%)}    \\ \hline
\multicolumn{2}{|c|}{Image Categories}                                                                                                                     \\ \hline
\multicolumn{1}{|c|}{a photo of a \{class label\}}                                                                                 & 64.9          \\ 
\multicolumn{1}{|c|}{how can you identify a \{class label\}}                                                                       & 61.7          \\ 
\multicolumn{1}{|c|}{A caption of an image of a \{class label\}}                                                                   & 64.5          \\ 
\multicolumn{1}{|c|}{A description of an image of a \{class label\}}                                                               & 64.7          \\ \hline
\multicolumn{2}{|c|}{Image Categories and Explanations}                                                                                                                      \\ \hline
\multicolumn{1}{|c|}{a photo of a \{class label\}. \{explanation\}}                                                                & 63.8          \\ 
\multicolumn{1}{|c|}{how can you identify a \{class label\}. \{explanation\}}                                                      & 65.5          \\ 
\multicolumn{1}{|c|}{A caption of an image of a \{class label\}. \{explanation\}}                                                  & 61.3          \\ 
\multicolumn{1}{|c|}{A description of an image of a \{class label\}. \{explanation\}}                                              & 63.5          \\ 
\multicolumn{1}{|c|}{a photo of a \{class label\}. Distinctive and physical features describing \{explanation\}}                   & 64.7          \\ 
\multicolumn{1}{|c|}{A caption of an image of a \{class label\}. Distinctive and physical features describing \{explanation\}}     & 62.1          \\ 
\multicolumn{1}{|c|}{A description of an image of a \{class label\}. Distinctive and physical features describing \{explanation\}} & 64.1          \\ 
\multicolumn{1}{|c|}{how can you identify a \{class label\}. Distinctive and physical features describing \{explanation\}}         & 66.0          \\ 
\multicolumn{1}{|c|}{how can you identify a \{class label\}. Distinctive and physical features describing \{explanation\} (E=2)}   & 66.3          \\ 
\multicolumn{1}{|c|}{how can you identify a \{class label\}. Distinctive and physical features describing \{explanation\} (E=3)}   & 66.5          \\ 
\multicolumn{1}{|c|}{how can you identify a \{class label\}. Distinctive and physical features describing \{explanation\} (E=4)}   & \textbf{66.6} \\ \hline
\end{tabular}
\label{prompt_zero_shot_transfer}
\end{table*}

\section{Additional Quantitative Evaluation}
In Table \ref{pretrain_unfiltered_results}, we present quantitative evaluation results on unfiltered scores for Uni-NLX, achieved through fine-tuning the pretrained captioning model of NLX-GPT. Our findings demonstrate that Uni-NLX achieves results comparable to NLX-GPT on VQA-X, ACT-X, and VQA-ParaX. Furthermore, it surpasses NLX-GPT performance on e-SNLI-VE and ImageNetX. Additionally, in the A-OKVQA task, our model outperforms NLX-GPT across three metrics, while also achieving comparable results on other metrics. In Table \ref{no_pretrain_filtered_results}, we provide results of Uni-NLX for the filtered setting without finetuning the pretrained captioning model. Our findings reveal that Uni-NLX surpasses NLX-GPT on ACT-X, e-SNLI-VE, and VCR, while achieving comparable results on the other tasks. By conducting a comparative analysis of both settings, it becomes evident that Uni-NLX exhibits superior performance in reasoning tasks such as e-SNLI-VE and VCR, as well as in visual recognition tasks such as ImageNetX and ACT-X.

\begin{table}
\caption{Unfiltered Scores for Uni-NLX compared to NLX-GPT on the 7 downstream tasks. Both models are w/ pretraining. B-N, M R, C, S are short for: BLEU-N, METEOR, ROUGE-L, CIDER and SPICE.}
\scalebox{0.85}{
\begin{tabular}{|c|cccccccc|}
\hline
        & \multicolumn{8}{c|}{VQA-X}                                                                                                                                                                                                                                                        \\ \hline
        & \multicolumn{1}{c|}{B1}            & \multicolumn{1}{c|}{B2}            & \multicolumn{1}{c|}{B3}            & \multicolumn{1}{c|}{B4}            & \multicolumn{1}{c|}{M}             & \multicolumn{1}{c|}{R}             & \multicolumn{1}{c|}{C}              & S             \\ \hline
NLX-GPT & \multicolumn{1}{c|}{\textbf{61.2}} & \multicolumn{1}{c|}{\textbf{46.1}} & \multicolumn{1}{c|}{\textbf{34.3}} & \multicolumn{1}{c|}{\textbf{25.6}} & \multicolumn{1}{c|}{\textbf{21.5}} & \multicolumn{1}{c|}{\textbf{48.7}} & \multicolumn{1}{c|}{\textbf{97.2}}  & \textbf{20.2} \\ 
Uni-NLX & \multicolumn{1}{c|}{60.2}          & \multicolumn{1}{c|}{44.7}          & \multicolumn{1}{c|}{32.8}          & \multicolumn{1}{c|}{24.1}          & \multicolumn{1}{c|}{20.8}          & \multicolumn{1}{c|}{47.2}          & \multicolumn{1}{c|}{89.9}           & 19.5          \\ \hline
        & \multicolumn{8}{c|}{ACT-X}                                                                                                                                                                                                                                                        \\ \hline
        & \multicolumn{1}{c|}{B1}            & \multicolumn{1}{c|}{B2}            & \multicolumn{1}{c|}{B3}            & \multicolumn{1}{c|}{B4}            & \multicolumn{1}{c|}{M}             & \multicolumn{1}{c|}{R}             & \multicolumn{1}{c|}{C}              & S             \\ \hline
NLX-GPT & \multicolumn{1}{c|}{\textbf{67.0}} & \multicolumn{1}{c|}{\textbf{50.5}} & \multicolumn{1}{c|}{\textbf{37.5}} & \multicolumn{1}{c|}{\textbf{28.1}} & \multicolumn{1}{c|}{\textbf{22.6}} & \multicolumn{1}{c|}{\textbf{49.7}} & \multicolumn{1}{c|}{\textbf{74.9}}  & \textbf{17.6} \\ 
Uni-NLX & \multicolumn{1}{c|}{66.6}          & \multicolumn{1}{c|}{\textbf{50.5}} & \multicolumn{1}{c|}{37.3}          & \multicolumn{1}{c|}{27.7}          & \multicolumn{1}{c|}{22.4}          & \multicolumn{1}{c|}{49.3}          & \multicolumn{1}{c|}{72.5}           & 17.2          \\ \hline
        & \multicolumn{8}{c|}{e-SNLI-VE}                                                                                                                                                                                                                                                    \\ \hline
        & \multicolumn{1}{c|}{B1}            & \multicolumn{1}{c|}{B2}            & \multicolumn{1}{c|}{B3}            & \multicolumn{1}{c|}{B4}            & \multicolumn{1}{c|}{M}             & \multicolumn{1}{c|}{R}             & \multicolumn{1}{c|}{C}              & S             \\ \hline
NLX-GPT & \multicolumn{1}{c|}{\textbf{34.3}} & \multicolumn{1}{c|}{\textbf{22.7}} & \multicolumn{1}{c|}{15.6}          & \multicolumn{1}{c|}{10.9}          & \multicolumn{1}{c|}{\textbf{17.5}} & \multicolumn{1}{c|}{31.7}          & \multicolumn{1}{c|}{106.6}          & \textbf{31.5} \\ 
Uni-NLX & \multicolumn{1}{c|}{33.9}          & \multicolumn{1}{c|}{\textbf{22.7}} & \multicolumn{1}{c|}{\textbf{15.8}} & \multicolumn{1}{c|}{\textbf{11.3}} & \multicolumn{1}{c|}{\textbf{17.5}} & \multicolumn{1}{c|}{\textbf{32.1}} & \multicolumn{1}{c|}{\textbf{107.5}} & \textbf{31.5} \\ \hline
        & \multicolumn{8}{c|}{VQA-ParaX}                                                                                                                                                                                                                                                    \\ \hline
        & \multicolumn{1}{c|}{B1}            & \multicolumn{1}{c|}{B2}            & \multicolumn{1}{c|}{B3}            & \multicolumn{1}{c|}{B4}            & \multicolumn{1}{c|}{M}             & \multicolumn{1}{c|}{R}             & \multicolumn{1}{c|}{C}              & S             \\ \hline
NLX-GPT & \multicolumn{1}{c|}{\textbf{37.9}} & \multicolumn{1}{c|}{\textbf{28.0}} & \multicolumn{1}{c|}{\textbf{21.5}} & \multicolumn{1}{c|}{\textbf{16.6}} & \multicolumn{1}{c|}{\textbf{19.5}} & \multicolumn{1}{c|}{\textbf{42.5}} & \multicolumn{1}{c|}{\textbf{156.6}} & \textbf{34.0} \\
Uni-NLX & \multicolumn{1}{c|}{36.8}          & \multicolumn{1}{c|}{27.2}          & \multicolumn{1}{c|}{20.8}          & \multicolumn{1}{c|}{16.1}          & \multicolumn{1}{c|}{19.1}          & \multicolumn{1}{c|}{42.0}          & \multicolumn{1}{c|}{152.6}          & 33.5          \\ \hline
        & \multicolumn{8}{c|}{A-OKVQA}                                                                                                                                                                                                                                                      \\ \hline
        & \multicolumn{1}{c|}{B1}            & \multicolumn{1}{c|}{B2}            & \multicolumn{1}{c|}{B3}            & \multicolumn{1}{c|}{B4}            & \multicolumn{1}{c|}{M}             & \multicolumn{1}{c|}{R}             & \multicolumn{1}{c|}{C}              & S             \\ \hline
NLX-GPT & \multicolumn{1}{c|}{57.1}          & \multicolumn{1}{c|}{\textbf{41.1}} & \multicolumn{1}{c|}{\textbf{30.4}} & \multicolumn{1}{c|}{\textbf{21.7}} & \multicolumn{1}{c|}{17.4}          & \multicolumn{1}{c|}{\textbf{46.8}} & \multicolumn{1}{c|}{\textbf{69.0}}  & 16.0          \\ 
Uni-NLX & \multicolumn{1}{c|}{\textbf{58.6}} & \multicolumn{1}{c|}{40.2}          & \multicolumn{1}{c|}{28.2}          & \multicolumn{1}{c|}{18.9}          & \multicolumn{1}{c|}{\textbf{17.5}} & \multicolumn{1}{c|}{44.8}          & \multicolumn{1}{c|}{61.1}           & \textbf{16.9} \\ \hline
        & \multicolumn{8}{c|}{ImageNetX}                                                                                                                                                                                                                                                    \\ \hline
        & \multicolumn{1}{c|}{B1}            & \multicolumn{1}{c|}{B2}            & \multicolumn{1}{c|}{B3}            & \multicolumn{1}{c|}{B4}            & \multicolumn{1}{c|}{M}             & \multicolumn{1}{c|}{R}             & \multicolumn{1}{c|}{C}              & S             \\ \hline
NLX-GPT & \multicolumn{1}{c|}{61.7}          & \multicolumn{1}{c|}{45.2}          & \multicolumn{1}{c|}{34.2}          & \multicolumn{1}{c|}{26.4}          & \multicolumn{1}{c|}{20.7}          & \multicolumn{1}{c|}{37.6}          & \multicolumn{1}{c|}{76.4}           & 20.2          \\ 
Uni-NLX & \multicolumn{1}{c|}{\textbf{63.2}} & \multicolumn{1}{c|}{\textbf{47.0}} & \multicolumn{1}{c|}{\textbf{36.0}} & \multicolumn{1}{c|}{\textbf{28.2}} & \multicolumn{1}{c|}{\textbf{21.4}} & \multicolumn{1}{c|}{\textbf{38.9}} & \multicolumn{1}{c|}{\textbf{82.8}}  & \textbf{21.1} \\ \hline
        & \multicolumn{8}{c|}{VCR}                                                                                                                                                                                                                                                          \\ \hline
        & \multicolumn{1}{c|}{B1}            & \multicolumn{1}{c|}{B2}            & \multicolumn{1}{c|}{B3}            & \multicolumn{1}{c|}{B4}            & \multicolumn{1}{c|}{M}             & \multicolumn{1}{c|}{R}             & \multicolumn{1}{c|}{C}              & S             \\ \hline
NLX-GPT & \multicolumn{1}{c|}{-}             & \multicolumn{1}{c|}{-}             & \multicolumn{1}{c|}{-}             & \multicolumn{1}{c|}{-}             & \multicolumn{1}{c|}{-}             & \multicolumn{1}{c|}{-}             & \multicolumn{1}{c|}{-}              & -             \\ 
Uni-NLX & \multicolumn{1}{c|}{19.1}          & \multicolumn{1}{c|}{10.1}          & \multicolumn{1}{c|}{5.8}           & \multicolumn{1}{c|}{3.6}           & \multicolumn{1}{c|}{9.1}           & \multicolumn{1}{c|}{20.0}          & \multicolumn{1}{c|}{24.9}           & 12.5          \\ \hline
\end{tabular}
}
\label{pretrain_unfiltered_results}
\end{table}

\begin{table}
\caption{Filtered Scores for Uni-NLX compared to NLX-GPT on the 7 downstream tasks. Both models are w/o pretraining.}
\scalebox{0.85}{
\begin{tabular}{|c|cccccccc|}
\hline
        & \multicolumn{8}{c|}{VQA-X}                                                                                                                                                                                                                                                        \\ \hline
        & \multicolumn{1}{c|}{B1}            & \multicolumn{1}{c|}{B2}            & \multicolumn{1}{c|}{B3}            & \multicolumn{1}{c|}{B4}            & \multicolumn{1}{c|}{M}             & \multicolumn{1}{c|}{R}             & \multicolumn{1}{c|}{C}              & S             \\ \hline
NLX-GPT & \multicolumn{1}{c|}{\textbf{63.3}} & \multicolumn{1}{c|}{\textbf{48.5}} & \multicolumn{1}{c|}{\textbf{36.9}} & \multicolumn{1}{c|}{\textbf{28.1}} & \multicolumn{1}{c|}{\textbf{22.6}} & \multicolumn{1}{c|}{\textbf{50.9}} & \multicolumn{1}{c|}{\textbf{108.5}} & \textbf{21.2} \\
Uni-NLX & \multicolumn{1}{c|}{60.3}          & \multicolumn{1}{c|}{45.0}          & \multicolumn{1}{c|}{33.0}          & \multicolumn{1}{c|}{24.2}          & \multicolumn{1}{c|}{20.7}          & \multicolumn{1}{c|}{48.2}          & \multicolumn{1}{c|}{91.9}           & 19.5          \\ \hline
        & \multicolumn{8}{c|}{ACT-X}                                                                                                                                                                                                                                                        \\ \hline
        & \multicolumn{1}{c|}{B1}            & \multicolumn{1}{c|}{B2}            & \multicolumn{1}{c|}{B3}            & \multicolumn{1}{c|}{B4}            & \multicolumn{1}{c|}{M}             & \multicolumn{1}{c|}{R}             & \multicolumn{1}{c|}{C}              & S             \\ \hline
NLX-GPT & \multicolumn{1}{c|}{69.5}          & \multicolumn{1}{c|}{53.5}          & \multicolumn{1}{c|}{40.7}          & \multicolumn{1}{c|}{31.3}          & \multicolumn{1}{c|}{24.8}          & \multicolumn{1}{c|}{52.3}          & \multicolumn{1}{c|}{99.6}           & 20.9          \\ 
Uni-NLX & \multicolumn{1}{c|}{\textbf{70.6}} & \multicolumn{1}{c|}{\textbf{55.6}} & \multicolumn{1}{c|}{\textbf{42.7}} & \multicolumn{1}{c|}{\textbf{32.9}} & \multicolumn{1}{c|}{\textbf{25.3}} & \multicolumn{1}{c|}{\textbf{52.9}} & \multicolumn{1}{c|}{\textbf{104.6}} & \textbf{22.4} \\ \hline
        & \multicolumn{8}{c|}{e-SNLI-VE}                                                                                                                                                                                                                                                    \\ \hline
        & \multicolumn{1}{c|}{B1}            & \multicolumn{1}{c|}{B2}            & \multicolumn{1}{c|}{B3}            & \multicolumn{1}{c|}{B4}            & \multicolumn{1}{c|}{M}             & \multicolumn{1}{c|}{R}             & \multicolumn{1}{c|}{C}              & S             \\ \hline
NLX-GPT & \multicolumn{1}{c|}{35.7}          & \multicolumn{1}{c|}{24.0}          & \multicolumn{1}{c|}{16.8}          & \multicolumn{1}{c|}{11.9}          & \multicolumn{1}{c|}{18.1}          & \multicolumn{1}{c|}{33.4}          & \multicolumn{1}{c|}{\textbf{114.7}} & \textbf{32.1} \\ 
Uni-NLX & \multicolumn{1}{c|}{\textbf{36.3}} & \multicolumn{1}{c|}{\textbf{24.8}} & \multicolumn{1}{c|}{\textbf{17.6}} & \multicolumn{1}{c|}{\textbf{12.8}} & \multicolumn{1}{c|}{\textbf{18.4}} & \multicolumn{1}{c|}{\textbf{33.8}} & \multicolumn{1}{c|}{114.6}          & 32.0          \\ \hline
        & \multicolumn{8}{c|}{VQA-ParaX}                                                                                                                                                                                                                                                    \\ \hline
        & \multicolumn{1}{c|}{B1}            & \multicolumn{1}{c|}{B2}            & \multicolumn{1}{c|}{B3}            & \multicolumn{1}{c|}{B4}            & \multicolumn{1}{c|}{M}             & \multicolumn{1}{c|}{R}             & \multicolumn{1}{c|}{C}              & S             \\ \hline
NLX-GPT & \multicolumn{1}{c|}{\textbf{40.7}} & \multicolumn{1}{c|}{\textbf{30.0}} & \multicolumn{1}{c|}{\textbf{23.2}} & \multicolumn{1}{c|}{\textbf{18.2}} & \multicolumn{1}{c|}{\textbf{21.1}} & \multicolumn{1}{c|}{\textbf{45.6}} & \multicolumn{1}{c|}{\textbf{187.5}} & \textbf{40.1} \\ 
Uni-NLX & \multicolumn{1}{c|}{38.5}          & \multicolumn{1}{c|}{28.4}          & \multicolumn{1}{c|}{21.7}          & \multicolumn{1}{c|}{17.0}          & \multicolumn{1}{c|}{20.6}          & \multicolumn{1}{c|}{44.9}          & \multicolumn{1}{c|}{179.8}          & 39.6          \\ \hline
        & \multicolumn{8}{c|}{A-OKVQA}                                                                                                                                                                                                                                                      \\ \hline
        & \multicolumn{1}{c|}{B1}            & \multicolumn{1}{c|}{B2}            & \multicolumn{1}{c|}{B3}            & \multicolumn{1}{c|}{B4}            & \multicolumn{1}{c|}{M}             & \multicolumn{1}{c|}{R}             & \multicolumn{1}{c|}{C}              & S             \\ \hline
NLX-GPT & \multicolumn{1}{c|}{\textbf{61.1}} & \multicolumn{1}{c|}{\textbf{45.7}} & \multicolumn{1}{c|}{\textbf{34.8}} & \multicolumn{1}{c|}{\textbf{26.1}} & \multicolumn{1}{c|}{\textbf{19.9}} & \multicolumn{1}{c|}{\textbf{51.0}} & \multicolumn{1}{c|}{\textbf{89.1}}  & \textbf{19.6} \\
Uni-NLX & \multicolumn{1}{c|}{62.8}          & \multicolumn{1}{c|}{43.6}          & \multicolumn{1}{c|}{30.7}          & \multicolumn{1}{c|}{21.3}          & \multicolumn{1}{c|}{19.1}          & \multicolumn{1}{c|}{47.9}          & \multicolumn{1}{c|}{73.5}           & 19.4          \\ \hline
        & \multicolumn{8}{c|}{ImageNetX}                                                                                                                                                                                                                                                    \\ \hline
        & \multicolumn{1}{c|}{B1}            & \multicolumn{1}{c|}{B2}            & \multicolumn{1}{c|}{B3}            & \multicolumn{1}{c|}{B4}            & \multicolumn{1}{c|}{M}             & \multicolumn{1}{c|}{R}             & \multicolumn{1}{c|}{C}              & S             \\ \hline
NLX-GPT & \multicolumn{1}{c|}{\textbf{72.3}} & \multicolumn{1}{c|}{\textbf{56.7}} & \multicolumn{1}{c|}{\textbf{45.0}} & \multicolumn{1}{c|}{\textbf{36.2}} & \multicolumn{1}{c|}{\textbf{26.0}} & \multicolumn{1}{c|}{44.6}          & \multicolumn{1}{c|}{\textbf{119.1}} & \textbf{28.1} \\ 
Uni-NLX & \multicolumn{1}{c|}{71.4}          & \multicolumn{1}{c|}{55.9}          & \multicolumn{1}{c|}{44.3}          & \multicolumn{1}{c|}{35.6}          & \multicolumn{1}{c|}{25.8}          & \multicolumn{1}{c|}{\textbf{44.7}} & \multicolumn{1}{c|}{117.7}          & 27.9          \\ \hline
        & \multicolumn{8}{c|}{VCR}                                                                                                                                                                                                                                                          \\ \hline
        & \multicolumn{1}{c|}{B1}            & \multicolumn{1}{c|}{B2}            & \multicolumn{1}{c|}{B3}            & \multicolumn{1}{c|}{B4}            & \multicolumn{1}{c|}{M}             & \multicolumn{1}{c|}{R}             & \multicolumn{1}{c|}{C}              & S             \\ \hline
NLX-GPT & \multicolumn{1}{c|}{24.7}          & \multicolumn{1}{c|}{15.0}          & \multicolumn{1}{c|}{9.6}           & \multicolumn{1}{c|}{6.6}           & \multicolumn{1}{c|}{12.2}          & \multicolumn{1}{c|}{26.4}          & \multicolumn{1}{c|}{46.9}           & 18.8          \\
Uni-NLX & \multicolumn{1}{c|}{\textbf{25.3}} & \multicolumn{1}{c|}{\textbf{19.6}} & \multicolumn{1}{c|}{\textbf{16.1}} & \multicolumn{1}{c|}{\textbf{14.0}} & \multicolumn{1}{c|}{\textbf{14.9}} & \multicolumn{1}{c|}{\textbf{30.5}} & \multicolumn{1}{c|}{\textbf{66.1}}  & \textbf{20.5} \\ \hline
\end{tabular}
}
\label{no_pretrain_filtered_results}
\end{table}

\section{Limitations and Collapse of VQA tasks}
\begin{figure}
    \centering
    \includegraphics[width=0.5\textwidth]{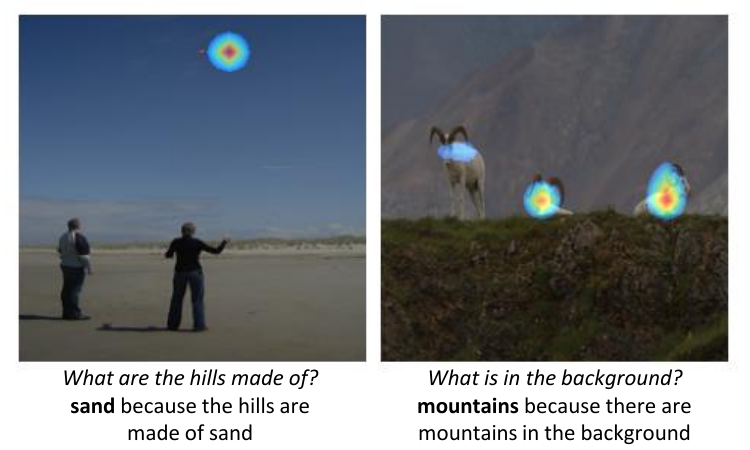}
    \caption{An example of the shortcut learning problem}
    \label{shortcut_learning}
\end{figure}

Acknowledging the limitations of our proposed model is of crucial importance. In particular, we address the issue of \textit{shortcut learning} that arises in explanations for some VQA tasks including A-OKVQA and VQA-ParaX. Specifically, the model generates an explanation which is composed of the answer and the question itself. For instance, consider the question "what is on the table?" and the predicted answer "cake", the generated explanation would be "there is cake on the table". Similarly, when presented with the question "what is on top of the plate?" and the predicted answer "pizza", the generated explanation would be "there is pizza on top of a plate". By using this shortcut approach, the model fails to reason correctly about the generated answer. This phenomenon becomes further evident by examining Figure \ref{shortcut_learning}, wherein we conduct an analysis of the heatmaps for various questions of the same images shown in Figure \ref{heatmap_answers}. As illustrated, the heatmaps exhibit no distinctions from those depicted in Figure \ref{heatmap_answers}, thereby indicating a lack of reasoning capabilities in generating explanations. Consequently, this finding corroborates the presence of the shortcut learning problem in the model. This problem is also observed in the individual uni-task models for A-OKVQA and VQA-ParaX. We intend to explore this matter in future research.

\end{document}